\documentclass[onecolumn]{article}

\usepackage[marginparwidth=75pt]{geometry}
\usepackage{times}
\usepackage{algorithm}	
\usepackage{algpseudocode}
\usepackage{algcompatible}

\def\1{\bm{1}}

\DeclareMathAlphabet{\mathsfit}{\encodingdefault}{\sfdefault}{m}{sl}
\SetMathAlphabet{\mathsfit}{bold}{\encodingdefault}{\sfdefault}{bx}{n}

\RequirePackage{natbib}		
\usepackage{hyperref}
\usepackage{amsmath}
\usepackage{amssymb}
\usepackage{mathtools}
\usepackage{amsthm}
\usepackage{thmtools,thm-restate}
\usepackage{bbm}
\usepackage{booktabs}
\usepackage{natbib}
\usepackage{amssymb}
\usepackage{mathtools}
\usepackage{amsthm}
\usepackage{arydshln} 
\theoremstyle{plain}
\newtheorem{theorem}{Theorem}

\theoremstyle{definition}

\newtheorem{assumption}{Assumption}

\theoremstyle{remark}
\newtheorem{remark}{Remark}

\usepackage{xcolor}

\def\fn{\footnote}

\usepackage{hyperref}
\usepackage{cleveref}  
\usepackage{xr}
\begin{document}

\title{Partially Functional Dynamic Backdoor Diffusion-based Causal Model}
\author{
    Xinwen Liu\\
    YunNan University\\
    \texttt{liuxinwen@stu.ynu.edu.cn} 
    \and
     Lei Qian\\
     Peking University\\
    \texttt{leiqian@stu.pku.edu.cn}
    \and
    Song Xi Chen\\
    Tsinghua University\\
    \texttt{sxchen@tsinghua.edu.cn}\and
    Niansheng Tang \\
    YunNan University\\
    \texttt{nstang@ynu.edu.cn}     
}
\date{}
\maketitle

\newcommand{\fix}{\marginpar{FIX}}
\newcommand{\new}{\marginpar{NEW}}

\begin{abstract}
Causal inference in spatio-temporal settings is critically hindered by unmeasured confounders with complex spatio-temporal dynamics and the prevalence of multi-resolution data. While diffusion models present a promising avenue for estimating structural causal models, existing approaches are limited by assumptions of causal sufficiency or static confounding, failing to capture the region-specific, temporally dependent nature of real-world latent variables or to directly handle functional variables. We bridge this gap by introducing the Partially Functional Dynamic Backdoor Diffusion-based Causal Model (PFD-BDCM), a unified generative framework designed to simultaneously tackle causal inference with dynamic confounding and functional data.  Our approach formalizes a novel structural causal model that captures spatio-temporal dependencies in latent confounders through conditional autoregressive processes, represents functional variables via basis expansion coefficients treated as standard graph nodes, and integrates valid backdoor adjustment into a diffusion-based generative process. We provide theoretical guarantees on the preservation of causal effects under basis expansion and derive error bounds for counterfactual estimates.  Experiments on synthetic data and a real-world air pollution case study demonstrate  that PFD-BDCM outperforms existing methods across observational, interventional, and counterfactual queries. This work provides a rigorous and practical tool for robust causal inference in complex spatio-temporal systems characterized by non-stationarity and multi-resolution data. 
    
\end{abstract}

\maketitle

\section{Introduction}\label{sec-intro}

Causal inference fundamentally addresses interventional 
and counterfactual questions that go beyond statistical correlations, proving valuable in high-stakes domains like healthcare for treatment effect estimation \citep{hill2011bayesian}, policy evaluation without randomized trials \citep{lalonde1986evaluating}. The field's core challenge stems from the fundamental problem of causal inference: the impossibility of simultaneously observing both an outcome under a treatment and the potential outcomes under the alternative treatment (or control)  for the same unit \citep{imbens2015causal}, necessitating methods to overcome confounding bias in observational data. 
Traditional approaches, including potential outcomes frameworks \citep{imbens2015causal}, propensity scoring \citep{ROSENBAUM198341}, instrumental variables \citep{ANGRIST1996444}, and structural causal models (SCMs) \citep{pearl2009causal}, exhibit significant limitations when handling modern complex datasets, as they struggle with high-dimensional confounders, ethical constraints of randomized trials, scarcity of valid instruments, and the requirement of known causal graphs. These limitations become particularly acute when confounders involve high-dimensional data like medical images or partially unobserved variables \citep{shalit2017estimating}.

Within the Structural Causal Model (SCM) framework, causal queries can be answered by learning a proxy for the unobserved exogenous noise and the structural equations \citep{pearl2009causal}. This suggests that (conditional) generative models that encode to a latent space could be an option for modeling SCMs, as the latent space serves as proxies for exogenous noises. Recent advances have explored the integration of deep generative models with structural causal models to address these challenges.  
Causal Normalizing Flows (CNFs) \citep{khemakhem2020causal} employ invertible transformations but face limitations in high-dimensional settings. Diffusion-based SCMs (Diff-SCM) \citep{mamaghan2023diffusion}  and Decoupled Flows (DecaFlow) \citep{almodovar2025decaflow} represent recent advances but lack explicit handling of spatio-temporal unmeasured confounding. 
 Diffusion-based Causal Models (DCM) \citep{chao2023interventional} and their extension incorporating backdoor adjustment (BDCM) \citep{shimizu2023} can learn complex structural equations from data.  
Nevertheless, both methods rely on static assumptions and thus do not fully capture the spatio-temporal structure inherent in confounding variables, a key characteristic of real-world systems, where such factors often demonstrate complex dependencies.
  
 To address this gap, we propose the Partially Functional Dynamic Backdoor Diffusion-based Causal Model (PFD-BDCM), a unified generative framework designed to simultaneously tackle causal inference with dynamic confounding and functional data.  
 We first formalize the Partially Functional Spatio-Temporal Dynamic SCM (PFST-DSCM), which represents unmeasured confounders via dynamic equations capturing spatial heterogeneity and temporal dependence, while incorporating functional variables through basis expansion, treating their coefficients as standard nodes in the causal graph. This formalization enables precise problem specification previously unattainable.  
 Building on this,  we develop Partially Functional Dynamic Backdoor Diffusion-based Causal Model (PFD-BDCM) for learning and inference through the Denoising Diffusion Implicit Models with Additional Conditioning (AC-DDIM). 
Unlike prior diffusion causal models (DCM  and  BDCM)  that condition only on parent variables  or a static backdoor adjustment set, AC-DDIM learns the distribution of a node given both its parents and a valid backdoor adjustment set, designed to directly adjust for bias from dynamically evolving unobserved variables. 
During training, the denoising network for each node receives the parent variables, backdoor adjustment variables and the noisy node value, conditioning the generative process to simulate removal of confounding even from unmeasured dynamic sources. For functional nodes, diffusion operates seamlessly on their finite-dimensional coefficient vectors. 
We provide theoretical guarantees for causal effect preservation under basis expansion and error bounds for counterfactual estimates. 
Through extensive simulations and a real-world case study on China's air pollution (2015-2020), we demonstrate PFD-BDCM's superior performance and practical utility for policy-relevant causal queries.

The remainder of this paper is organized as follows. Section~\ref{sec:SN}  introduces necessary notations and reviews necessary background, Section \ref{sec:PFD-BDCMs} provides the PFST-DSCM and the PFD-BDCM architecture and algorithms. Section \ref{sec:theory} presents our theoretical guarantees, which prove that causal effects  are preserved under basis expansion and derive explicit error bounds linking the model's counterfactual estimation error to its empirical reconstruction fidelity, providing a rigorous foundation for the proposed methodology. 
In Section~\ref{sec:SS}, we report on extensive simulation studies that evaluate PFD‑BDCM across a variety of data‑generating mechanisms, highlighting its advantages in handling dynamic confounding and functional data. Section~\ref{sec:Application} illustrates the practical utility of the framework through a real-world case study on atmospheric pollutant emissions in China (2015-2020). The paper concludes in Section~\ref{sec:Concluding} with a summary of contributions, a discussion of limitations, and an outline of directions for future research.

\section{Notation and Fundamental}\label{sec:SN}
We first introduce some useful notations, fundamental concepts of causal inference.

\textbf{Notations.} Let $[n] := {1, \dots, n}$ index spatial units (regions) and $[J] := {1, \dots, J}$ index time points,  $[K] := {1, \dots, K}$ index nodes in a directed acyclic graph (DAG) $\mathcal{G}$. For node $k$, denote its parent set by $\mathbb{PA}_k$ and a valid backdoor adjustment set by $\mathbb{B}_k$. The corresponding endogenous variables are $\mathbf{x}_{\mathbb{PA}_k}$ and $\mathbf{x}_{\mathbb{B}_k}$. Each node is associated with an endogenous variable $\textnormal{x}_k \in \mathbb{R}^{d_k}$ and an exogenous noise $\textnormal{u}_k \in \mathbb{R}^{d_k}$. For functional nodes, $d_k = K_n$, representing the number of basis functions. And for scalar nodes, $d_k=1$. In our diffusion process, $\hat{\textnormal{u}}_k^t$ and $\hat{\textnormal{x}}_k^t$ denote the latent and reconstructed variables at step $t$.

In causal inference, a confounder denotes a variable that causally influences both a treatment variable $\textnormal{x}$ and an outcome variable $\textnormal{y}$, thereby inducing a non-causal association between them.  
 Observable Confounders refer to confounders that can be measured,  which permit adjustment through statistical methods such as stratification, matching or regression  \citep{pearl2009causal}. Unobservable confounders denote latent variables that fulfill confounding criteria but resist direct measurement, which can potentially bias causal estimates when unaccounted for.   Unobserved explanatory (explained) nodes are unobserved confounder nodes that have no  nodes (unobserved confounder nodes)  as its parent (descendant). 
 
\begin{figure}[!ht]
    \includegraphics[width=0.5\textwidth]{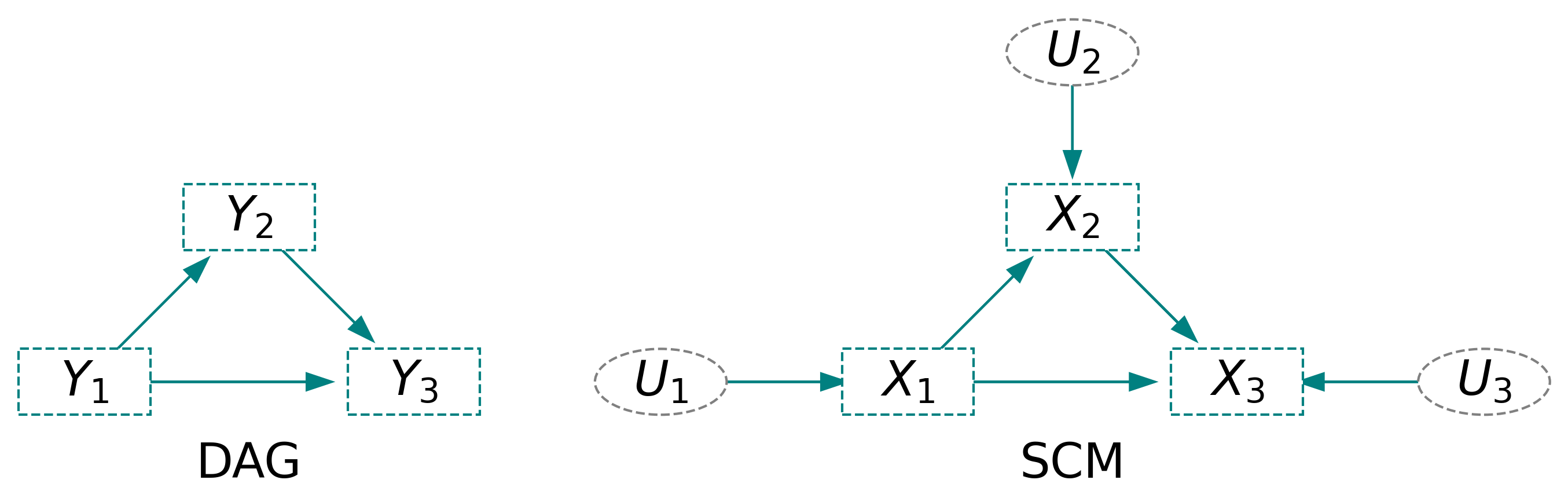}
    \centering
    \caption{DAG with three nodes (left) and SCM with three exogenous and endogenous nodes (right)}
    \label{SCM}
\end{figure}

\noindent \textbf{Structural Causal Models.}
Consider a DAG (such as Fig. \ref{SCM}) $\mathcal{G}$ with nodes $[K]$  in a topologically sorted order   
\fn{A topological order is a linear arrangement of variables where a variable appears after all its direct causes (parents)  \citep{pearl2009causal}.}, where a node $k$ is built with an endogenous random variable $\textnormal{x}_k$ defined on a space $\mathcal{X}_k \subset \mathbb{R}^{d_k}$, which has a random exogenous input $\textnormal{u}_k$. 
A structural causal model (SCM) $\mathcal{M}$ {characterizes the relationship between the endogenous variable $\textnormal{x}_k$  of a node $k$ with the endogenous variables of its parents $\mathbf{x}_{\mathbb{PA}_k}$ and its own exogenous variable $\textnormal{u}_k$.}  Formally, we define $\mathcal{M}:=(\mathbf{f}(\mathbf{x},\mathbf{u}), p_\mathbf{u})$,  where $\mathbf{f}(\mathbf{x},\mathbf{u})$ specifies how entire endogenous variables $\mathbf{x}:=\{\textnormal{x}_1, \cdots, \textnormal{x}_K\}$ are generated from the set of exogenous random variables $\mathbf{u}:=\{\textnormal{u}_1, \cdots, \textnormal{u}_K\}$ with a prior distribution $p_\mathbf{u}$. 
The structural mechanism is governed by $\mathbf{f}(\mathbf{x},\mathbf{u}):=(f(\mathbf{x}_{\mathbb{PA}_1},\textnormal{u}_1), \cdots, f(\mathbf{x}_{\mathbb{PA}_K},\textnormal{u}_K))$, where each $\textnormal{x}_k:=f(\mathbf{x}_{\mathbb{PA}_k},\textnormal{u}_k)$ for $k\in[K]$~\citep{pearl2009causal}.

\section{Methods}\label{sec:PFD-BDCMs}

We begin by formalizing the data-generating process underlying complex spatio‑temporal data through a Spatio‑Temporal Dynamic Structural Causal Model (ST‑DSCM). Building on this foundation, we then introduce the Partially Functional Dynamic Backdoor Diffusion‑based Causal Model (PFD‑BDCM), which is designed to answer causal queries under the extended PFST‑DSCM framework.

\subsection{PFST-DSCM}\label{sec: PFST-DSCM} 

Consider a spatio-temporal dataset $\mathbf{x}:= \{\textnormal{x}_k\}_{k \in [K]}$ containing $K$ variables across $n$ regions over $J$ time points. Using $i$ to index regions and $j$ for time points,  $ \textnormal{x}_k = (x_{k,ij})_{n \times J} $ represent the value of the $k$-th variable $ \textnormal{x}_k $ over the time points and regions. We use a DAG to characterize the causal relationships among $\{\textnormal{x}_k\}_{k \in [K]}$.  Consider a DAG $\mathcal{G}$ with nodes $[K]$ and a topologically sorted order such that each node $k$ has the  $\textnormal{x}_k$ as the random variable.  
Let $\mathbb{C}_1$ and $\mathbb{C}_2 \subseteq [K]$ be two distinct sets of nodes with unobserved confounders, where  $\mathbb{C}_1$ designates a set of unobserved explanatory nodes and $\mathbb{C}_2$ denotes a set of unobserved explained nodes.  For $h=1, 2$, let $\mathbf{x}_{\mathbb{C}_h}:=\{\mathbf{x}_{{\mathbb{C}_h}, i j}\}_{i \in [n], j \in [J]}:=\{x_{q, i j}\}_{i \in [n], j \in [J], q \in \mathbb{C}_h}$ and $\mathbf{u}_{\mathbb{C}_h}:=\{\mathbf{u}_{{\mathbb{C}_h}, i j}\}_{i \in [n], j \in [J]}:=\{u_{q, i j}\}_{i \in [n], j \in [J], q \in \mathbb{C}_h}$. 

By the definition of $\mathbb{C}_1$, every node in this set satisfies the backdoor criterion\fn{A set of node $\mathbb{B}$ satisfies backdoor criterion \citep{pearl2016} for tuple $(\textnormal{x}, \textnormal{y})$ in DAG $\mathcal{G}$ if no node in $\mathbb{B}$ is a descendant of $\textnormal{x}$ and $\mathbb{B}$ blocks all paths between $\textnormal{x}$ (cause) and $\textnormal{y}$ (outcome) which contains an arrow into $\textnormal{x}$.}. 
For nodes with unmeasured confounders ($k \in \mathbb{C}_2$, $\mathbb{PA}_k \in \mathbb{C}_1$), we assume spatio-temporal dynamic SCM relations:
\begin{equation}\label{eq:dscm}
\textnormal{x}_{k, ij} = f_{k, i j}(\mathbf{x}_{{\mathbb{PA}_k},ij}, \textnormal{u}_{k,ij}), 
\end{equation}
where $f_{k,ij}$ represents a class of potentially linear or nonlinear spatio-temporal functions, and $\textnormal{u}_{k,ij}$ denotes independent exogenous noise terms. 

Specifically, to account for spatial heterogeneity, we posit a relationship such that 
\begin{equation}\label{eq:spatial heterogeneity}
\mathbf{x}_{{\mathbb{C}_2}, i j}=\boldsymbol{\Gamma}_i \boldsymbol{G}
\left(\mathbf{x}_{{\mathbb{C}_1}, i j}\right)
+\mathbf{u}_{{\mathbb{C}_2}, i j},
\end{equation}
where $\boldsymbol{\Gamma}_i$ is a $\dim(\mathbf{x}_{{\mathbb{C}_2}, i j}) \times \dim(\mathbf{x}_{{\mathbb{C}_1}, i j})$ structural coefficient matrix varying across regions; 
 $\boldsymbol{G}(\cdot)=(G_1(.), \cdots, G_q(.))^\top$ is a $q \times 1$ nonzero  vector-valued function with differentiable functions $G_1,\cdots, G_q$, and  $q = \dim(\mathbf{x}_{{\mathbb{C}_1}, i j})$.  
 
To capture the temporal dependence,  let $\mathbf{x}_{{\mathbb{C}_1}, i}=(\mathbf{x}_{{\mathbb{C}_1}, i 1}^\top, \cdots, \mathbf{x}_{{\mathbb{C}_1}, i J}^\top)^\top$ and $\mathbf{u}_{{\mathbb{C}_2}, i}=(\mathbf{u}_{{\mathbb{C}_2}, i 1}^\top, \cdots, \mathbf{u}_{{\mathbb{C}_2}, i J}^\top)^\top$,  the covariance matrices  
$\mathrm{Cov}(\mathbf{x}_{{\mathbb{C}_1}, i})$  and  $\mathrm{Cov}(\mathbf{u}_{{\mathbb{C}_2}, i})$ of $\mathbf{x}_{{\mathbb{C}_1}, i}$ and $\mathbf{u}_{{\mathbb{C}_2}, i}$ can be expressed as
\begin{equation}\label{eq7}
\mathrm{Cov}(\mathbf{x}_{{\mathbb{C}_1}, i})=\boldsymbol{D}_{{\mathbb{C}_1} ,i}
\otimes 
\mathrm{Cov}(\mathbf{x}_{{\mathbb{C}_1,ij}}),\quad
\mathrm{Cov}(\mathbf{u}_{{\mathbb{C}_2}, i})=\boldsymbol{D}_{{\mathbb{C}_2}, i}
\otimes 
\mathrm{Cov}(\mathbf{u}_{{\mathbb{C}_2},ij}),
\end{equation}
where $\boldsymbol{D}_{{\mathbb{C}_h}, i} (h \in \{1,2\})$ are the $J \times J$ adjacent time covariance matrices, $\mathrm{Cov}(\cdot)$ represent between-variable covariances. In order to establish a rigorous framework for the temporal adjacency structure, the conditional autoregressive (CAR) model \citep{Besag1974SpatialIA} is adopted with the forms
\begin{equation}\label{CAR}
\begin{aligned}
\boldsymbol{D}_{{{\mathbb{C}_h}, i}}
=
\left(\boldsymbol{I}_{J}-\rho_{{{\mathbb{C}_h}, i}} \boldsymbol{H}_{{{\mathbb{C}_h}, i}}\right)^{-1},\quad
\end{aligned}
\end{equation}
where $\rho_{{{\mathbb{C}_h}, i}} (h \in \{1,2\})$ are adjacent time association parameters, and $\boldsymbol{H}_{{{\mathbb{C}_h}, i}} (h \in \{1,2\})$ 
are $J \times J$ adjacency matrices  in which the element $h_{j l}=1$ implies that time $l$ is adjacent to time $j$ and otherwise $h_{j l}=0$.  
The covariance kernel $\boldsymbol{D}_{\mathbb{C}_1, i}$ explicitly models temporal dependency by inducing correlation across neighboring time points. This represents a fundamental departure from the standard independent and identically distributed (i.i.d.) assumption typically placed on latent variables in conventional causal generative models, allowing our framework to capture the realistic, evolving nature of unmeasured confounders.

For other endogenous variables, we assume standard SCM relations 
\begin{equation}\label{eq:scm}
\textnormal{x}_{k,ij} = f_k(\mathbf{x}_{{\mathbb{PA}_k},ij}, \textnormal{u}_{k,ij}), 
\end{equation}
where $f_k$ are potentially linear (or nonlinear) static functions. 

We refer to the model satisfying the above specifications as the Spatio-Temporal Dynamic Structural Causal Model (ST-DSCM). It provides a principled framework for capturing spatio-temporally dynamic unmeasured confounding.  
However, Real-world applications often present significant challenges in elucidating causal relationships among variables observed under heterogeneity. To address these complexities, we extend the ST-DSCM by incorporating functional random variables $\textnormal{z}_{ij}(t)$. 
We adopt a basis  expansion framework to achieve dimensionality reduction in the functional space via a set of orthogonal basis functions
$\{\mathbf{b}_m(t)\}_{m=1}^{K_n}$. 
Define 
\begin{equation}\label{eq:basis expansion}
\textnormal{x}_{m,ij} = \int b_{m}(t) \textnormal{z}_{ij}(t) \ \mathrm{d}t, 
\end{equation}
for $i \in [n], j \in [J], m \in [K_n]$. Let $\mathbf{x}_{k,ij} = (\textnormal{x}_{k,1,ij}, \dots, \textnormal{x}_{k, K_n,ij})^\top$ denote the collection of all basis coefficients derived from a functional variable $\textnormal{z}_{ij}(t)$. 
This technique projects infinite-dimensional functional covariates onto a finite-dimensional space while preserving functional characteristics. The mathematical details of the basis expansion and its properties are provided in Supplementary \ref{sec: FDR}.  
Crucially, for causal inference, this coefficient vector $\mathbf{x}_{k,ij}$ is treated as a single multidimensional endogenous node within the causal graph. This representation allows all standard causal operations (intervention, counterfactual reasoning, and estimation) to be applied uniformly, whether the node originates from a scalar or a functional measurement. 

By augmenting the ST-DSCM with this representation, we obtain the Partially Functional Spatio-Temporal Dynamic Structural Causal Model (PFST-DSCM) \fn{For example, Fig.~\ref{DAG for PFST-DSCMs model} is a PFST-DSCM with 11 exogenous and endogenous nodes (where nodes $\textnormal{x}_{3,ij}$, $\textnormal{x}_{4,ij}$ and $\textnormal{x}_{9,ij}$ are unmeasured confounder nodes with spatial heterogeneity and temporal dependencies, $\textnormal{z}_{ij}(t) $ is functional nodes, and $\mathbf{x}_{2,ij}$ is the corresponding base expansion vector node)}. 
In this model, nodes can be scalar, dynamic latent confounders (as in the ST-DSCM), or finite-dimensional vectors representing functional variables. This extension preserves the dynamic modeling of unmeasured confounding while seamlessly integrating multi-resolution (scalar and functional) data.

\subsection{Identification of the PFST-DSCM}\label{sec:Identification}

In causal inference, defining the model as a Directed Acyclic Graph (DAG) (Assumption \ref{ass:PFST-DDAG}(i)) is a standard and necessary foundation \citep{pearl2009causal}. The designation ``Partially Functional Spatio-Temporal Dynamic'' accurately captures the specific composition of our graph $\mathcal{G}$, which integrates standard scalar nodes, dynamically modeled latent confounder nodes, and multi-dimensional nodes representing functional data via basis coefficients. This precise nomenclature is essential to distinguish our framework from conventional DAGs or those used in static causal models like DCM or BDCM.  Crucially, our model identification relies on a set of core assumptions (Assumption  \ref{ass:CAR Model} and Assumption \ref{ass:PFST-DDAG}((ii), (iii)).  A detailed discussion justifying the plausibility of these assumptions  is provided in the Supplementary \ref{sec: PFST-DSCM-sup}. 

\begin{assumption}\label{ass:CAR Model}
For CAR model (Eq.(\ref{CAR})),  $\rho_{\mathbb{C}_h i} $  must satisfy the following constrained conditions $\rho_{\mathbb{C}_h i} ^L<\rho_{\mathbb{C}_h i} <\rho_{\mathbb{C}_h i} ^U$ to ensure that  $\boldsymbol{D}_{\mathbb{C}_h i}$ are positive definite, where $\rho_{\mathbb{C}_h i} ^L=\min (1 /\tau_{\mathbb{C}_h  J}, 0), \rho_{\mathbb{C}_h i} ^U=1/\tau_{\mathbb{C}_h  1}$, and $\tau_{\mathbb{C}_h  1} \geq \cdots \geq \tau_{\mathbb{C}_h  J}$  are the eigenvalues of neighborhood matrices $\boldsymbol{H}_{\mathbb{C}_h  i}$  \citep{song2012bayesian}.
\end{assumption}
\begin{assumption}\label{ass:PFST-DDAG}
(i) The Partially Functional Spatio-Temporal Dynamic Directed Acyclic Graph (PFST-DDAG) $\mathcal{G}$ is the graph induced by PFST-DSCM $\mathcal{M}$.  

(ii) The unobserved random variables are jointly independent (Markovian SCM), and every PFST-DSCM $\mathcal{M}$ entails a unique joint observational distribution satisfying the causal Markov assumption: $q(\mathbf{x}) = \prod_{k=1}^K q(\textnormal{x}_k| \mathbf{x}_{\mathbb{PA}_k})$.

(iii) For any observed node with unobserved parents, a valid backdoor adjustment set $\mathbb{B}_k$ can be identified from $\mathcal{G}$, enabling causal identification even in the presence of dynamic unmeasured confounders under Assumption \ref{ass:CAR Model}.
\end{assumption}

The PFST-DSCM provides a unified formalization that captures our motivating air pollution setting: confounders that evolve in space and time, and drivers that may be functional in nature. The model operates under the standard assumption that the causal graph is known (or estimable), and all structural parameters can be estimated from data using existing Bayesian spatio-temporal methods \citep{song2012bayesian, TangBayesian2017}. With this formalization in place, the framework supports the full range of causal queries (observational, interventional, and counterfactual) from observational data.

\subsection{PFD-BDCM}\label{sec: PFD-BDCM}

We now present the PFD-BDCM, a model designed to handle causal queries under the PFST-DSCM framework, which explicitly accounts for unmeasured confounders with spatial heterogeneity and temporal dependence. The model employs an encoder-decoder architecture to enable causal reasoning across multi-resolution variables.

\subsubsection{Data-Generating}\label{sec: DGP}

The data-generating process of the PFD-BDCM is formalized as follows. 
i) If node \(k\) and its parent nodes are unobservable confounders, the data \(\mathbf{x}_k \in \mathbb{R}^{d_k}\) 
are generated via Eq. \eqref{eq:dscm}. These nodes are not themselves targets of training, their data serve only to provide structural information and to adjust for bias during the training of observable nodes of interest. Consequently, no diffusion model is trained on data from such nodes.  
ii) For any other node \(k\), data are generated using Eq. \eqref{eq:scm}.   When node \(k\) and all its parent nodes are observable, an encoder \(g_k\) maps \((\mathbf{x}_{\mathbb{PA}_k}, \mathbf{x}_k)\) to a latent variable \(\widehat{\mathbf{u}}_k := g_k(\mathbf{x}_{\mathbb{PA}_k}, \mathbf{x}_k)\), which captures the information of the exogenous noise \(\mathbf{u}_k\). A decoder \(h_k\) then reconstructs \(\mathbf{x}_k\) as \(\widehat{\mathbf{x}}_k := h_k(\mathbf{x}_{\mathbb{PA}_k}, \widehat{\mathbf{u}}_k)\). This setup corresponds to the data-generating process of the DCM \citep{chao2023interventional}.  And when node \(k\) is observable but some of its parent nodes are unobservable confounders, we introduce a backdoor adjustment set \(\mathbb{B}_k\) for node \(k\) as a proxy for the unobserved parents. 
The encoder \(g_k\) then maps \((\mathbf{x}_{\mathbb{PA}_k}, \mathbf{x}_{\mathbb{B}_k}, \mathbf{x}_k)\) to \(\widehat{\mathbf{u}}_k := g_k(\mathbf{x}_{\mathbb{PA}_k}, \mathbf{x}_{\mathbb{B}_k}, \mathbf{x}_k)\), and the decoder \(h_k\) reconstructs \(\mathbf{x}_k\) as \(\widehat{\mathbf{x}}_k := h_k(\mathbf{x}_{\mathbb{PA}_k}, \mathbf{x}_{\mathbb{B}_k}, \widehat{\mathbf{u}}_k)\).  Conditioning on the backdoor adjustment set \(\mathbf{x}_{\mathbb{B}_k}\) provides a mechanism to account for confounding.

\subsubsection{AC-DDIM}\label{sec: AC-DDIM}

Our approach  leverages diffusion models to learn the structural equations of a causal system, representing each endogenous variable $\textnormal{x}_k$ with a dedicated conditional diffusion model. For counterfactual reasoning, we require a deterministic mapping between observations and latent codes. Denoising Diffusion Implicit Models (DDIMs) \citep{ddim} provide such a deterministic non-Markovian reverse process, essential for identifiability. 
While standard conditional diffusion models condition only on parent variables $\mathbf{x}_{\mathbb{PA}_k}$, and existing backdoor-adjusted models condition on a backdoor set $\mathbf{x}_{\mathbb{B}_k}$, AC-DDIM conditions on the combined set $\mathbf{x}_{\mathbb{PB}_k} = \mathbf{x}_{\mathbb{PA}_k} \cup \mathbf{x}_{\mathbb{B}_k}$. This ensures the generative process for $\textnormal{x}_k$ is conditioned on the proper adjustment set, accounting for dynamically generated unmeasured confounders. During training, the denoising network learns to reconstruct $\textnormal{x}_k$ given $\mathbf{x}_{\mathbb{PB}_k}$; during sampling, providing values for $\mathbf{x}_{\mathbb{PB}_k}$ yields statistically adjusted samples.

Formally, for each node $k \in [K]$, the latent variable $\hat{\textnormal{u}}_k := \textnormal{x}_k^T$ is generated through the forward implicit diffusion process 
\begin{equation}
\begin{aligned} \label{Enc-BDCM}
\textnormal{x}^{t}_k :=& \sqrt{1-\bar\alpha_{t} / (1-\bar\alpha_{t-1})} \textnormal{x}^{t-1}_k + \left( \sqrt{\bar\alpha_{t}} - \sqrt{\bar\alpha_{t-1}(1-\bar\alpha_t-\sigma^2_t)/(1-\bar\alpha_{t-1}} \right)\textnormal{x}_k^0\\
&+\sigma_t\epsilon_k,\quad \epsilon_k \sim \mathcal{N}(0, \mathbf{I}),\text{for} \quad t \in[T].
\end{aligned}
\end{equation}
This latent representation $\hat{\textnormal{u}}_k$ serves as a proxy for the exogenous noise $\textnormal{u}_k$.

The reconstruction $\hat{\textnormal{x}}_k := \hat{\textnormal{x}}_k^0$ is obtained via the reverse implicit diffusion process 
\begin{equation}
\begin{aligned} \label{Dec-BDCM}
\hat{\textnormal{x}}^{t-1}_k := &\sqrt{ 1/\alpha_t} \hat{\textnormal{x}}^t_k -  \left( \sqrt{(1-\bar\alpha_{t})/\alpha_t} - \sqrt{1-\bar\alpha_{t-1}} \right)\epsilon_{k, \theta}^t( \hat{\textnormal{x}}_k^t, \mathbf{x}_{\mathbb{PB}_k}),
\end{aligned}
\end{equation}
for $t = T,\cdots,1$, initialized with $\hat{\textnormal{x}}^T_k := \hat{\textnormal{u}}_k$. Since AC-DDIM's generative process is non-Markovian and deterministic (when $\sigma_t=0$), timesteps can be skipped for acceleration (see Supplementary \ref{sec: AC-DDIM-sup}).

The parameters $\theta$ are learned by minimizing the following loss function 
\begin{equation}\label{eq:loss}
\begin{aligned}
      L_\gamma(\epsilon_{k, \theta}) & := \sum_{\tau_i=\tau_S}^{\tau_1}\gamma_{\tau_i}\mathbb{E}_{{\tau_i}, \textnormal{x}_k^0, \mathbf{x}_{\mathbb{PB}_k}, \epsilon_k}\left[  \left\|\epsilon_{k, \theta}^{\tau_i}\left( \sqrt{\bar\alpha_{\tau_i}} \textnormal{x}_k^0 + \sqrt{1 -\bar \alpha_{\tau_i}} \epsilon_k,  \mathbf{x}_{\mathbb{PB}_k}\right) - \epsilon_k\right\|_2^2 \right],
\end{aligned}
\end{equation}
where $\gamma := [\gamma_{\tau_s}, \ldots, \gamma_{\tau_1}]$ is a vector of positive weighting coefficients that depends on $\alpha_{\tau_S:\tau_1}$.

For functional nodes represented via basis expansion, the diffusion process operates on the finite-dimensional coefficient vectors in $\mathbb{R}^{K_n}$. The derivation of AC-DDIM follows the standard DDIM \citep{ddim} framework with the extension of conditioning on additional covariates, complete mathematical formulations and theoretical details are provided in Supplementary \ref{sec: AC-DDIM-sup}.  

The encoding and decoding functions for node $k$ are denoted as $\mathrm{Enc}_k(\mathbf{x}_{\mathbb{PB}_k},\textnormal{x}^0_k )$ (Eq. \eqref{Enc-BDCM}) and $\mathrm{Dec}_k( \mathbf{x}_{\mathbb{PB}_k},\hat{\textnormal{u}}_k)$ (Eq.\eqref{Dec-BDCM}) respectively, the pseudocodes are provided in Supplementary  \ref{Enc&Dec} (Algorithm \ref{algo: PFD-BDCM encoding} and \ref{algo: PFD-BDCM decoding} ). 
    
\subsubsection{Training PFD-BDCMs  and Answering Causal Queries}\label{sec:Training}

 The comprehensive training methodology
 (Algorithm \ref{algo:PFD-BDCMs-training})  incorporates backdoor adjustment sets as covariates while training distinct diffusion models per scale node  $\textnormal{x}_{k}$ (or functional node vector  $\mathbf{x}_{k}$ ). Crucially, generative models for endogenous nodes exhibit mutual independence during training, thereby enabling parallelized optimization. This parallelism is feasible since each diffusion model necessitates only its target node's values and corresponding backdoor adjustment set values. The final PFD-BDCM architecture integrates these $K$ trained diffusion models $\{\epsilon_{k, \theta}\}_{k \in [K]}$.

\begin{algorithm}[!ht]
    \caption{ \textbf{PFD-BDCMs Training} }
    \label{algo:PFD-BDCMs-training}
    \textbf{Input:} Observational dataset $\mathcal{D} = \{\mathbf{x}_{ij}\}_{i\in[n],j\in[J]}$, causal graph $\mathcal{G}$ (PFST-DDAG) with $K$ nodes $\{\mathbf{x}_k\}_{k\in[K]}$, noise schedule $\{\bar{\alpha}_t\}_{t=1}^{T}$, subsequence $\{\tau_i\}_{i=1}^{S}$ (optional, for accelerated training), learning rate $\eta$.
    
    \textbf{Output:} Trained denoising networks $\{\epsilon_{k,\theta}\}_{k\in[K]}$.
    \begin{algorithmic}[1]
        \STATE Initialize parameters $\theta_k$ for each node $k \in [K]$.
        \WHILE{not converged}
        \STATE Sample a mini-batch $\{\mathbf{x}^{(b)}\}_{b=1}^{B} \sim \mathcal{D}$.
         \FOR{$k=1$ to $K$} 
            \STATE Extract target node values: $\mathbf{x}_k \leftarrow \{\mathbf{x}_k^{(b)}\}_{b=1}^{B}$.
            \STATE Extract conditioning set values: $\mathbf{x}_{\mathbb{PB}_k} \leftarrow \{\mathbf{x}_{\mathbb{PB}_k}^{(b)}\}_{b=1}^{B}$.
            \STATE Sample a diffusion time step $t \sim \mathrm{Uniform}(\{1, \dots, T\})$ (or $\tau \sim \mathrm{Uniform}(\{\tau_1, \dots, \tau_S\})$ if using subsequence).
            \FOR{$b=1$ to $B$}
                \STATE Sample noise: $\boldsymbol{\epsilon}^{(b)} \sim \mathcal{N}_{d_k}(\mathbf{0}, \mathbf{I})$.
                \STATE Compute noisy input: $\tilde{\mathbf{x}}_k^{(b)} \leftarrow \sqrt{\bar{\alpha}_t} \mathbf{x}_k^{(b)} + \sqrt{1 - \bar{\alpha}_t} \boldsymbol{\epsilon}^{(b)}$.
            \ENDFOR
            \STATE Compute loss: $\mathcal{L}_k \leftarrow \frac{1}{B} \sum_{b=1}^{B} \| \boldsymbol{\epsilon}^{(b)} - \epsilon_{k,\theta}( \tilde{\mathbf{x}}_k^{(b)}, t, \mathbf{x}_{\mathbb{PB}_k}^{(b)} ) \|_2^2$.
            \STATE Update parameters: $\theta_k \leftarrow \theta_k - \eta \nabla_{\theta_k} \mathcal{L}_k$.
        \ENDFOR
        \ENDWHILE
    \STATE Return trained denoising networks $\{\epsilon_{k,\theta}\}_{k\in[K]}$.
    \end{algorithmic}
\end{algorithm}

We now elucidate the methodology for leveraging trained PFD-BDCMs to approximate diverse causal queries. Resolution of observational and interventional queries necessitates sampling from their respective observational and interventional distributions. Counterfactual queries, however, operate at unit granularity by modifying structural equation assignments while preserving the latent exogenous noise variables consistent with empirical observations. 

\noindent \textbf{Generating samples for observational/interventional queries.}
Interventional queries concern the causal effect of actively setting a variable to a specific value, which is formally represented by the $do$-operator,  as in $p_\theta(\textnormal{x}|\mathrm{do}(\textnormal{x}_\mathcal{L}:=\boldsymbol{\gamma}))$ .
Specifically, to generate samples approximating the interventional distribution  $p_\theta(\textnormal{x}_k|\hat{\mathbf{x}}_{\mathbb{PB}_k,-\mathcal{L}},\mathrm{do}(\mathbf{x}_\mathcal{L}:=\boldsymbol{\gamma}))$ using a trained PFD-BDCM model, 
 where $\hat{\mathbf{x}}_{\mathbb{PB}_k,-\mathcal{L}}$ denotes $\hat{\mathbf{x}}_{\mathbb{PB}_k}$ with the intervention node variables $\textnormal{x}_\mathcal{L}$ removed.  We implement the following procedure: i) For intervened node $l \in \mathcal{L}$,  we set $\hat{\textnormal{x}}_l:= \gamma_l$ deterministically;  ii) For root nodes $k$, sample $\hat{\textnormal{x}}_k$ from empirical training distributions;  iii) For non-intervened nodes $k \notin \mathcal{L}$, sample latent variable $\hat{\textnormal{u}}_k \sim \mathcal{N}(0, 1)$  ( or $\hat{\textnormal{u}}_k \sim t_\nu(0, 1)$,  where  \(t_\nu(\boldsymbol{\mu}, \boldsymbol{\Sigma})\) is a multivariate t‑distribution with \(\nu\) degrees of freedom, location vector \(\boldsymbol{\mu}\), and scale matrix \(\boldsymbol{\Sigma}\) ).  And subsequently compute $\hat{\textnormal{x}}_k:=\mathrm{Dec}_k(\hat{\mathbf{x}}_{\mathbb{PB}_k}, \hat{\textnormal{u}}_k )$  utilizing inductively generated parent variable values and backdoor  adjustment set values. Generated values propagate to descendant nodes as parent  and backdoor adjustment inputs. Specially, observational sampling  $p_\theta(\textnormal{x}_k|\mathbf{x}_{\mathbb{PB}_k})$ corresponds to $\mathcal{L} = \emptyset$.
The pseudocode  formalized in Supplementary \ref{ACQ} (Algorithm~\ref{algo: PFD-BDCM intervention sampling}). 

\noindent \textbf{Counterfactual Queries} 
concern hypothetical scenarios given actual observed outcomes, and require three steps: \textit{abduction} (inferring exogenous noise), \textit{action} (modifying equations), and \textit{prediction} (simulating new outcomes).
Specifically, to compute counterfactual estimates $\hat{\boldsymbol{x}}^\mathrm{CF}:= (\boldsymbol{x}_1^\mathrm{CF} ,\cdots,\boldsymbol{x}_K^\mathrm{CF} )$ , within the PFD-BDCM framework, given factual observation $\boldsymbol{x}^\mathrm{F}  := (\boldsymbol{x}_1^\mathrm{F} ,\cdots,\boldsymbol{x}_K^\mathrm{F} )$ ,  and intervention set $\mathcal{L}$ with values $\boldsymbol{\gamma}$, we implement the following systematic procedure: 
i) For intervened nodes $l \in \mathcal{L}$, assign $\hat{\boldsymbol{x}}_l^\mathrm{CF} := \boldsymbol{\gamma}_l$  deterministically; ii) For non-intervened descendant nodes $k$, using inductively generated parent and backdoor adjustment estimates $\hat{\boldsymbol{x}}_{\mathbb{PB}_k}^\mathrm{CF}$, we compute $\hat{\boldsymbol{x}}_k^\mathrm{CF} := \mathrm{Dec}_k ( \hat{\boldsymbol{x}}_{\mathbb{PB}_k}^\mathrm{CF},\mathrm{Enc}_k(\boldsymbol{x}_{\mathbb{PB}_k}^\mathrm{F},\boldsymbol{x}_k^\mathrm{F}))$ ,  where factual noise is implicitly encoded. 
The complete formalization appears in 
Supplementary \ref{ACQ} (
Algorithm \ref{algo: PFD-BDCM counterfactual inference}).

\section{Theoretical Guarantees}\label{sec:theory}

This section develops the theoretical underpinnings for integrating functional data into structural causal models via basis expansion, establishing rigorous guarantees for the counterfactual estimation accuracy of the proposed PFD-BDCM framework. 

\noindent \textbf{Notation:} We consider an endogenous variable $\textnormal{x}_k$ ( or  $ \mathbf{x}^{(z)}$ for functional nodes ) governed by structural equation $\textnormal{x}_k := f_{k}(\mathbf{x}_{\mathbb{PB}_k}, \textnormal{u}_k)$ ( or $\mathbf{x}_k^{(z)} = f_k^{(z)}(\mathbf{x}_{\mathbb{PB}_k^{(z)}}, \mathbf{u}_k^{(z)})$) with parent set and backdoor adjustment set $\mathbf{x}_{\mathbb{PB}_k}$ ( or $\mathbf{x}_{\mathbb{PB}_k^{(z)}}$) and exogenous noise $\textnormal{u}_k$ ( or  $\mathbf{u}_k^{(z)}$). We analyze a single node without loss of generality (by permutation invariance of nodes), henceforth denoting the target variable as $\textnormal{x} \in \mathcal{X} \subseteq \mathbb{R}$ ( or the  target vector $ \mathbf{x}^{(z)} \in \mathcal{X}^{K_n} \subseteq  \mathbb{R}^{K_n}$), its parents and  backdoor adjustments as $\mathbf{x}_{\mathbb{PB}} \in \mathcal{X}_{\mathbb{PB}} \subseteq \mathbb{R}^K$ ( or $\mathbf{x}_{\mathbb{PB}^{(z)}} \in \mathcal{X}_{\mathbb{PB}^{(z)}} \subseteq \mathbb{R}^{K}$), and exogenous noise as $\textnormal{u}$ ( or $ \mathbf{u}^{(z)}$). The encoder-decoder architecture comprises
\begin{align*}
&g : \mathcal{X}\times \mathcal{X}_{\mathbb{PB}} \to \mathcal{U} \quad( \text{or}\quad  g^{(z)} : \mathcal{X}^{K_n} \times \mathcal{X}_{\mathbb{PB}}   \to \mathcal{U}^{K_n} )\quad \text{(encoding function)};\\ 
&h : \mathcal{U} \times \mathcal{X}_{\mathbb{PB}}  \to \mathcal{X} \quad (\text{or}\quad  h^{(z)} : \mathcal{U}^{K_n} \times \mathcal{X}_{\mathbb{PB}}  \to \mathcal{X}^{K_n} ) \quad\text{(decoding function)},
\end{align*}
where $\mathcal{U}$ ( or $\mathcal{U}^{K_n}$) denotes the (${K_n}$ dimensional ) latent space. Within PFD-BDCM, $g$ ( or  $g^{(z)}$) and $h$ ( or $h^{(z)}$) correspond to the $\mathrm{Enc}(\cdot,\cdot)$ and $\mathrm{Dec}(\cdot,\cdot)$ operators, respectively. 

\subsection{Approximate Error Bound of Functional Data}\label{sec:Theoretical CI-FD}

First, a fundamental challenge is ensuring that the finite-dimensional basis approximation of $z(t)$ preserves its true causal effect.
Theorem  \ref{thm:preservation}  formally addresses this concern. Under standard smoothness assumption  (Assumption \ref{ass:smoothness}) , we establish that the interventional distribution of any outcome under $\operatorname{do}(z(t)=z_0(t))$ converges weakly to the distribution under $\operatorname{do}(\mathbf{x}^{(z)}=\mathbf{x}^{(z)}_0)$ as the basis dimension $K_n$ increases.  Detailed proofs are provided in the Supplementary \ref{Theorem-proof-func}.

\begin{assumption}
\label{ass:smoothness}
     The coefficient function $c(t)$ and the functional random variable  $\textnormal{z}(t)$ belong to a Hilbert space $\mathcal{H} \subseteq L^2(\mathcal{T})$ with inner product $\langle \cdot, \cdot \rangle_\mathcal{H}$ and norm $\|\cdot\|_\mathcal{H}$, and the orthonormal basis $\{b_m(t)\}_{m=1}^\infty$ is complete in $\mathcal{H}$ . Moreover, there exist  constants $s_c, C_c,s_z, C_z > 0$ such that the approximation error for any $K_n \in \mathbb{N}$, 
     $$
    \left\| c(t) - \sum_{m=1}^{K_n} \phi_m b_m(t) \right\|_{\mathcal{H}} \leq C_c K_n^{-s_c},
    \qquad
    \mathbb{E} \left[ \left\| \textnormal{z}(t) - \sum_{m=1}^{K_n} \textnormal{x}_{m} b_m(t) \right\|_{\mathcal{H}}^2 \right] \leq C_z K_n^{-2s_z},
    $$
    where $\phi_m = \langle c(t)), b_m(t) \rangle_\mathcal{H}$ and $\textnormal{x}_m = \langle \textnormal{z}(t), b_m(t) \rangle_\mathcal{H}$ are the basis coefficients.
\end{assumption}

Assumption~\ref{ass:smoothness} is standard in functional data analysis and nonparametric estimation. First, the requirement that the function space $\mathcal{H}$ possesses a complete orthonormal basis guarantees that any element, here the coefficient function $c(t)$ and the realizations of the functional covariate $z(t)$, can be represented as a linear combination of the basis functions, thereby justifying the use of projection‑based or sieve estimation approaches. Second, the specific power‑law decay rates $K_n^{-s_c}$ and $K_n^{-s_z}$ for the truncation errors quantify the smoothness of the underlying functions; functions with higher smoothness (i.e., larger $s_c$, $s_z$) can be approximated more accurately with a finite number of basis terms. These rates are crucial for establishing the theoretical convergence rates of the resulting estimators. Common examples satisfying such conditions include functions in Sobolev or Hölder spaces when Fourier, Spline, or Wavelet bases are used. The moment condition on $z(t)$ ensures that the approximation error holds in a mean‑squared sense across its distribution.

\begin{theorem}
\label{thm:preservation}
    Under Assumption~\ref{ass:smoothness}, the causal effect of the functional variable $\textnormal{z}(t)$ on any endogenous variable $\textnormal{x}$ in the infinite-dimensional SCM is equivalent to the causal effect of the finite-dimensional basis coefficient vector $\mathbf{x}^{(z)}$ on  $\textnormal{x}$ in the PFST-DSCM, as $K_n \to \infty$. Specifically, for any intervention $\text{do}(\textnormal{z}(t) = z_0(t))$, the resulting interventional distribution on $\textnormal{x}$ converges weakly to the interventional distribution obtained by the intervention $\text{do}(\mathbf{x}^{(z)} = \boldsymbol{x}_0^{(z)})$ in the finite-dimensional model, where $\boldsymbol{x}_0^{(z)}$  is the projection of $z_0(t)$ onto the basis.
\end{theorem}

 Theorem \ref{thm:preservation}  implies that all statistical functionals (including means, variances, and quantiles) that are continuous with respect to the weak topology will converge. In particular, the causal effect of $\textnormal{z}(t)$ on $\textnormal{x}$, defined as a functional of the interventional distribution (e.g., the average treatment effect), is preserved in the limit. Thus, the finite-dimensional intervention  $\text{do}(\mathbf{x}^{(z)} = \boldsymbol{x}_0^{(z)})$ yields an interventional distribution that converges weakly to that of the infinite-dimensional intervention $\text{do}(\textnormal{z}(t) = z_0(t))$, and the causal effect of the functional variable $\textnormal{z}(t)$ is equivalently captured by the basis coefficient representation as the number of basis functions increases.

Next, Theorem \ref{thm:error_bound_Finite-Dimensional} establishes an upper bound on the approximation error induced by the finite-dimensional projection, which decays with the number of basis functions. Detailed proofs are provided in Supplementary \ref{Theorem-proof-func}.

\begin{theorem}
\label{thm:error_bound_Finite-Dimensional}
    Under Assumption~\ref{ass:smoothness}, assume further that the intervention function $z_0(t)$ satisfies a deterministic approximation error bound: there exist constants $C_{z_0}, s_{z_0} > 0$ such that for any $K_n \in \mathbb{N}$,
    $$
    \| z_0(t) - \sum_{m=1}^{K_n} x_{0,m} b_m(t) \|_{\mathcal{H}} \leq C_{z_0} K_n^{-s_{z_0}},
    $$
where $x_{0,m} = \langle z_0(t), b_m(t) \rangle_{\mathcal{H}}$ (assuming orthonormal basis). Let $B_c = \| c(t) \|_{\mathcal{H}}$ and $B_{z_0} = \| z_0(t) \|_{\mathcal{H}}$ be finite. Then, for any structural function $f$ that is Lipschitz continuous with respect to the functional term with Lipschitz constant $L_f$, and for any intervention $\text{do}(\textnormal{z}(t) = z_0(t))$, the pointwise error satisfies
  $$
\begin{aligned} 
   & \left\| f(\dots, \langle c(t), z_0(t) \rangle, \dots, \textnormal{u}) - f(\dots, \boldsymbol{\phi}^\top \boldsymbol{x}_0^{(z)}, \dots, \textnormal{u}) \right\| \\
    \leq & L_f \left( B_{z_0} C_c K_n^{-s_c} + (B_c + C_c) C_{z_0} K_n^{-s_{z_0}} \right).
    \end{aligned}
    $$
Consequently, the error is of order $O(K_n^{-\min(s_c, s_{z_0})})$. In particular, the error in the interventional mean of $\textnormal{x}$ (i.e., the average causal effect) decays at the same polynomial rate.
\end{theorem}

Theorem \ref{thm:error_bound_Finite-Dimensional} establishes the convergence rate for the approximation error incurred when representing infinite-dimensional functional quantities in causal estimands with a finite-dimensional basis expansion. The total pointwise error decomposes into terms arising from the approximation of the functional coefficient $c(t)$ and the intervention function $z_0(t)$, each decaying at a polynomial rate dictated by their respective smoothness parameters $s_c$ and $s_{z_0}$. Consequently, the overall error is of order $O(K_n^{-\min(s_c, s_{z_0})})$, demonstrating that smoother underlying functions permit accurate estimation with fewer basis functions. This result directly implies the same convergence rate for the average causal effect, providing a theoretical foundation for sieve estimation in functional causal inference by formally characterizing the trade-off between model complexity (governed by $K_n$) and approximation accuracy.

\subsection{Counterfactual Error Bounds}\label{Error Bounds}

The accuracy of any counterfactual query hinges on a model's ability to correctly perform \textit{abduction}, that is, to recover the exogenous noise that generated the observed data. This step is what differentiates causal models from purely predictive ones. Our theoretical analysis establishes that under our framework, this recovery is possible, and its fidelity directly controls the final counterfactual error. To comprehensively validate PFD-BDCM, we analyze three progressively complex settings: scalar nodes (the baseline case), functional (multidimensional) nodes, and finally, scenarios with heavy-tailed noise or model misspecification. This structure mirrors the increasing complexity of real-world data and allows us to pinpoint the assumptions required at each step.

\subsubsection{One-Dimensional Counterfactual Error Bounds}\label{Error Bounds-LD}

We first establish the core error bound for a scalar endogenous variable $\textnormal{x} \in \mathbb{R}$ ( detailed proofs show in Supplementary \ref{sup:Error Bounds-LD}). This forms the theoretical backbone, which we later extend. 
Our theoretical results rely on a set of assumptions regarding the structural equation and the encoder-decoder model. These conditions are essential for ensuring that the latent variable learned by the encoder can uniquely recover the unobserved exogenous noise, which is the cornerstone of accurate counterfactual estimation \citep{lu2020sample,nasr2023counterfactual}.
For a variable $\textnormal{x} \in \mathcal{X}\subset \mathbb{R}$ with structural equation $\textnormal{x}:=f(\mathbf{x}_\mathbb{PB}, \textnormal{u})$ 
where  $\textnormal{u} \perp \!\!\! \perp \mathbf{x}_\mathbb{PB}$, we have the following assumptions:

\begin{assumption}\label{ass:Encoder Independence}
The encoding function produces representations that are statistically independent of the parent variables and the backdoor adjustment variables: $g(\mathbf{x}_{\mathbb{PB}}, \textnormal{x}) \perp\!\!\!\perp \ \mathbf{x}_{\mathbb{PB}}$.
\end{assumption}
\begin{assumption}\label{ass:Encoder Invertibility}
The encoding function is invertible and differentiable with respect to the endogenous variable for fixed  parent variables and backdoor adjustment variables: $g(\mathbf{x}_{\mathbb{PB}},\cdot): \mathcal{X} \to \mathcal{U} \text{ is bijective and } C^1, \quad \text{for all }\mathbf{x}_{\mathbb{PB}} \in \mathcal{X}_{\mathbb{PB}}.$
\end{assumption}
\begin{assumption}\label{ass:Generalized Structural Regularity}
The structural equation is continuously differentiable in both arguments and strictly monotonic in the exogenous noise for each fixed value of the parent variables and the backdoor adjustment variables
$f \in C^1(\mathcal{X}_{\mathbb{PB}}\times \mathcal{U}, \mathcal{X}), \quad\frac{\partial f(\mathbf{x}_{\mathbb{PB}}, \textnormal{u})}{\partial \textnormal{u}} > 0 \quad \text{for all } \mathbf{x}_{\mathbb{PB}} \in \mathcal{X}_{\mathbb{PB}}, \textnormal{u} \in \mathcal{U}.$
\end{assumption}

These assumptions, while formal, are well-motivated in the context of causal inference and deep generative models.  
Assumption \ref{ass:Encoder Independence}  ensures that the encoder learns a pure representation of the exogenous noise uncontaminated by information from the backdoor adjustment variables. In practice, this can be enforced through regularization or architectural constraints. This is naturally satisfied in settings like additive noise models with $f(\mathbf{x}_\mathbb{PB},\textnormal{u}) = f^*(\mathbf{x}_\mathbb{PB})+\textnormal{u}$ where $\mathbf{x}_\mathbb{PB}$ and $\textnormal{u}$ is independent. If the fitted model $\hat{f}\equiv f^*$, then $g(\mathbf{x}_\mathbb{PB}, \textnormal{x}) =\textnormal{u}$.   
Assumption  \ref{ass:Encoder Invertibility} is intrinsically satisfied by the bijective properties of deterministic diffusion architectures \citep{ddim}, guaranteeing uniqueness in latent representations while preserving compatibility with standard implementations.  
Assumption \ref{ass:Generalized Structural Regularity} is satisfied by major identifiable model classes, including additive noise, post-nonlinear, and heteroscedastic formulations \citep{strobl2023identifying} while concurrently resolving symmetric noise ambiguities characteristic of observational data. This assumption further aligns with contemporary identifiability frameworks \citep{nasr2023counterfactual} and intrinsically precludes non-identifiable structural equations. Additional justifications  of this assumptions' reasonableness in practical settings are provided in Supplementary \ref{sup:Error Bounds-LD}.

These assumptions collectively ensure that the encoded latent $\hat{\textnormal{u}} = g(\mathbf{x}_{\mathbb{PB}}, \textnormal{x})$ is an invertible transform of the true noise $\textnormal{u}$ (Theorem \ref{theo:General Exogenous Noise Recovery}).

\begin{theorem}\label{theo:General Exogenous Noise Recovery}
Under Assumptions \ref{ass:Encoder Independence}, \ref{ass:Encoder Invertibility}, and \ref{ass:Generalized Structural Regularity}, the encoded latent variable $\hat{\textnormal{u}} = g(\mathbf{x}_{\mathbb{PB}}, \textnormal{x})$ is related to the true exogenous noise $\textnormal{u}$ through an invertible transformation. That is, there exists an invertible function $\phi: \mathcal{U} \to \mathcal{U}$ such that $\hat{\textnormal{u}} = \phi(\textnormal{u})$.
\end{theorem}

Theorem \ref{theo:General Exogenous Noise Recovery}  implies that the abduction step  $g(\mathbf{x}_\mathbb{PB}, \textnormal{x})$ correctly captures the essence of the unobserved exogenous random variable $\textnormal{u}$ that generated the factual observation. We now explore the direct consequences of this result. 
In an oracle scenario where the model achieves perfect reconstruction which means $h(\mathbf{x}_\mathbb{PB}, g(\mathbf{x}_\mathbb{PB}, \textnormal{x})) =\textnormal{x}$ holding almost surely (a.s.), the counterfactual estimate will be ``perfect''. This precise case implies the satisfaction of Theorem \ref{theo:General Exogenous Noise Recovery}, yielding the relationship $h( \mathbf{x}_\mathbb{PB}, \phi(\textnormal{u} )) = f(\mathbf{x}_\mathbb{PB}, \textnormal{u})$. Consequently, when making a counterfactual prediction for a new intervention $\mathbf{x}_\mathcal{L} := \boldsymbol{\gamma}$, the model computes $h(\mathbf{x}_\mathbb{PB,-\mathcal{L}}\cup \boldsymbol{\gamma}, \phi( \textnormal{u} ))$, which a.s. equates to the true counterfactual $f(\mathbf{x}_\mathbb{PB,-\mathcal{L}}\cup \boldsymbol{\gamma}, u)$.

In empirical settings, exact reconstruction is rarely achievable.
 From this foundation, we derive our central result (Theorem \ref{theo:General Counterfactual Error Bound}): for any factual observation and counterfactual intervention, the estimation error is bounded by the model's empirical reconstruction error.

\begin{theorem}\label{theo:General Counterfactual Error Bound}
Let \(\mathcal{X}\) be a metric space of functions with metric \(d: \mathcal{X} \times \mathcal{X} \to \mathbb{R}_{\ge 0}\) that induces the topological structure of interest (e.g., the \(L^2\) norm for square-integrable functions).  Under Assumptions \ref{ass:Encoder Independence}, \ref{ass:Encoder Invertibility}, and  \ref{ass:Generalized Structural Regularity},  for any factual observation $(x^F, \mathbf{x}_{\mathbb{PB}}^F)$ generated by $x^F = f(\mathbf{x}_{\mathbb{PB}}^F, u)$ and any counterfactual intervention $\mathrm{do}(\mathbf{x}_{\mathcal{L}}^{CF} := \boldsymbol{\gamma})$, if the reconstruction error is bounded by $\tau$ with respect to \(d\), i.e.,
$d(h( \mathbf{x}_{\mathbb{PB}}, g(\mathbf{x}_{\mathbb{PB}}, \textnormal{x})), \textnormal{x}) \leq \tau$, then the counterfactual estimation error is bounded by $\tau$: $
d(\hat{x}^{\mathrm{CF}}, x^{\mathrm{CF}}) \leq \tau.
$
\end{theorem}

This bound (Theorem \ref{theo:General Counterfactual Error Bound}) is powerful and elegant because it directly connects a standard, optimizable training objective (reconstruction loss) to the causal fidelity of the model. This provides a clear theoretical justification for using a powerful generative model (diffusion) for causal inference: improving its generative performance guarantees better counterfactual estimates. This link is not explicitly established in prior diffusion-based causal works like BDCM, which focus more on algorithmic feasibility than on such performance-based guarantees.

\subsubsection{Multidimensional Counterfactual Error Bound}\label{Error Bounds-HD}

Handling functional data constitutes a methodological component of PFD-BDCM. Theoretically, this entails generalizing the scalar assumptions to the multidimensional vector $\mathbf{x}^{(z)} \in \mathbb{R}^{K_n}$ of basis coefficients. Specifically, Assumptions \ref{ass:Encoder Independence}–\ref{ass:Generalized Structural Regularity} are extended by replacing scalar monotonic functions with \textit{diffeomorphisms} (invertible smooth maps), as detailed in Assumptions \ref{ass:multidim_struct}–\ref{ass:multidim_indep} in Supplementary \ref{sup:Error Bound-HD-proof}. This generalization preserves the principle of ``noise recovery'' in the higher-dimensional setting.

For counterfactual queries that involve both functional and scalar nodes, additional regularity of the causal paths is required. We distinguish three canonical cases, each requiring a specific cross‑dimensional Lipschitz condition. The details  show in Supplementary \ref{sup:Error Bound-HD-proof} (Assumption \ref{ass:case1}-\ref{ass:case3}).  Under these assumptions we obtain the following multidimensional counterpart of Theorem \ref{thm:multidim_error},  and all detailed proofs show in Supplementary \ref{sup:Error Bound-HD-proof}.

\begin{theorem}
\label{thm:multidim_error}
    Assume the relevant node‑type assumptions (Assumptions \ref{ass:multidim_struct}--\ref{ass:multidim_indep} for functional nodes, and Assumptions \ref{ass:Encoder Independence}--\ref{ass:Generalized Structural Regularity} for scalar nodes) hold. Then the counterfactual estimation error satisfies the following bounds for the three canonical cases:
    
    \textbf{Case 1} (functional intervention, scalar outcome): Under Assumption \ref{ass:case1},  $d(\hat{x}^{\mathrm{CF}} , x^{\mathrm{CF}}) \le \tau + L_{\text{cross}} \bigl(\tau^{(z)} + O(K_n^{-s})\bigr)$.
                            
  \textbf{Case 2}  (functional intervention, functional outcome): Under Assumption \ref{ass:case2},             $              d(\hat{\mathbf{x}}^{(z')^{\mathrm{CF}}} , \mathbf{x}^{(z')^{\mathrm{CF}}}) \le \tau^{(z')} + O(K_n^{-s}) + O(K_n'^{-s'})$.
                  
    \textbf{Case 3} (scalar intervention, functional outcome): Under Assumption \ref{ass:case3},               $d(\hat{\mathbf{x}}^{(z)^{\mathrm{CF}}} ,  \mathbf{x}^{(z)^{\mathrm{CF}}}) \le \tau^{(z)} + L_{\text{cross}}^{(z)} \tau_l + O(K_n^{-s})$.

Here $\tau$, $\tau^{(z)}$, $\tau^{(z')}$, $\tau_l$ denote the reconstruction errors of the corresponding nodes, measured in the appropriate norms (e.g., Euclidean norm for vectors, absolute value for scalars). The terms $O(K_n^{-s})$ and $O(K_n'^{-s'})$ are the basis‑expansion errors for the functional variables involved.
\end{theorem}

The multidimensional treatment has several practical advantages: i) It is computationally more efficient to train a single diffusion model for the whole vector of basis coefficients than separate models for each coefficient; ii) The joint model captures dependencies among the coefficients, leading to statistically more consistent estimates; iii) The approach scales gracefully with the number of basis functions provided the Lipschitz constants remain bounded.
In current research on causal diffusion models,  no existing work has systematically addressed how to incorporate functional data into the causal diffusion modeling framework with accompanying theoretical guarantees. Our multidimensional framework fills this gap. It is the first to enable the modeling of functional data within a causal diffusion model and provide a corresponding error analysis, offering a novel theoretical tool for handling multi-resolution spatio-temporal data.

\subsubsection{Robustness to Heavy-Tailed Noise and Model Misspecification}\label{sec:Robustness and Misspec.} 

To demonstrate the practical utility and robustness of our framework, we extend our theoretical analysis beyond idealized settings, considering realistic challenges such as heavy-tailed noise distributions and potential model misspecification. Detailed formulations of the corresponding assumptions, theorems, and complete proofs for these extended scenarios are provided in the Supplementary \ref{sec:HTN} and \ref{sec:lim}.

\textbf{Heavy-Tailed Noise.} Real-world environmental data often exhibit outliers. Under a polynomial tail assumption (Assumption \ref{Heavy-Tailed Exogenous Noise} in Supplementary \ref{sec:HTN} ) on the noise $\textnormal{u}$, we show (Theorem \ref{theo:Moment-Based Counterfactual Bound} in Supplementary \ref{sec:HTN} ) that the counterfactual error inherits a similar polynomial concentration bound. More importantly, we prove (Theorem \ref{theo:Robust Encoder Design} in Supplementary \ref{sec:HTN}  ) that using a robust loss function (e.g., Huber loss) for the encoder can recover an \textit{exponential} concentration rate, significantly tightening the error control.  

\begin{remark}
    Theorem \ref{theo:Moment-Based Counterfactual Bound} provides a polynomial concentration bound for counterfactual estimation errors when dealing with heavy-tailed noise distributions. This result demonstrates that the tail of the counterfactual error is controlled by the tail of the reconstruction error. Hence, by controlling the tail of the reconstruction error, we control the tail of the counterfactual error. Specifically, the tail behavior is governed by the worse of these two moment conditions, with constants $C_\tau$ depending on the Lipschitz properties and model parameters. This polynomial decay reflects the fundamental challenges posed by heavy-tailed noise in causal inference.
\end{remark}
\begin{remark}
   Theorem \ref{theo:Robust Encoder Design} shows that employing Huber loss in the encoder design achieves exponential concentration despite heavy-tailed noise. The robust M-estimation approach ensures that the encoder error probability decays exponentially fast as $\epsilon$ increases, providing significantly tighter control over estimation errors compared to the polynomial bound. This demonstrates the substantial benefits of robust estimation techniques in handling heavy-tailed distributions. 
\end{remark}

\textbf{Model Misspecification.} The theoretical guarantees established so far rely on the assumption that the structural equation follows an additive noise model (Assumption~\ref{ass: Additive Noise Model}) or, more generally, a strictly monotonic form (Assumption~\ref{ass:Generalized Structural Regularity}). In practice, the true data-generating process may deviate from these idealized forms. To address this, we now analyze the robustness of the PFD-BDCM framework when the structural equation is only approximately additive or when the encoder-decoder pair does not perfectly satisfy the invertibility and independence conditions. We discuss the limitations of our model assumptions (Supplementary \ref{sec:Assumption Violations and Robustness}) and provide error bounds under model misspecification, which are consolidated in Proposition \ref{prop:general-robustness} and its detailed proof all show in  Supplementary \ref{sec:Model Misspecification}. 

\begin{remark}
Proposition~\ref{prop:general-robustness} extends the error analysis to account for model misspecification. It establishes that the counterfactual error of PFD-BDCM remains bounded even when the true structural equations deviate from the assumed form, provided the deviation is controlled. The bound explicitly decomposes the total error into components arising from encoder inaccuracy ($\delta_n$), decoder approximation error ($\tau_n$), and the degree of misspecification ($\zeta$). Under heavy-tailed noise, employing a robust encoder yields near-exponential error concentration, with performance primarily limited by the misspecification term $\eta$. This result demonstrates that the framework maintains controlled error under model mismatch and remains applicable to real-world problems, such as the air pollution case study, where ideal assumptions are often violated.
\end{remark}

Our theoretical analysis provides a cohesive and multi-layered foundation for PFD-BDCM. These results do more than justify our specific algorithm; they advance the theoretical understanding of what makes deep generative causal inference work, particularly when dealing with the spatio-temporal and functional complexities that define modern environmental datasets.

\section{Simulation Studies}\label{sec:SS}

Our synthetic study served as a controlled setting to evaluate the contribution of each methodological innovation within PFD-BDCM. We assessed the performance of a total of 11 models, including our proposed PFD-BDCM, and a range of baseline and alternative approaches, across eight distinct structural equation scenarios. 

\subsection{Data Generation}\label{sec:DG}
\begin{figure}[!ht]
\centering
\includegraphics[width=0.7\textwidth]{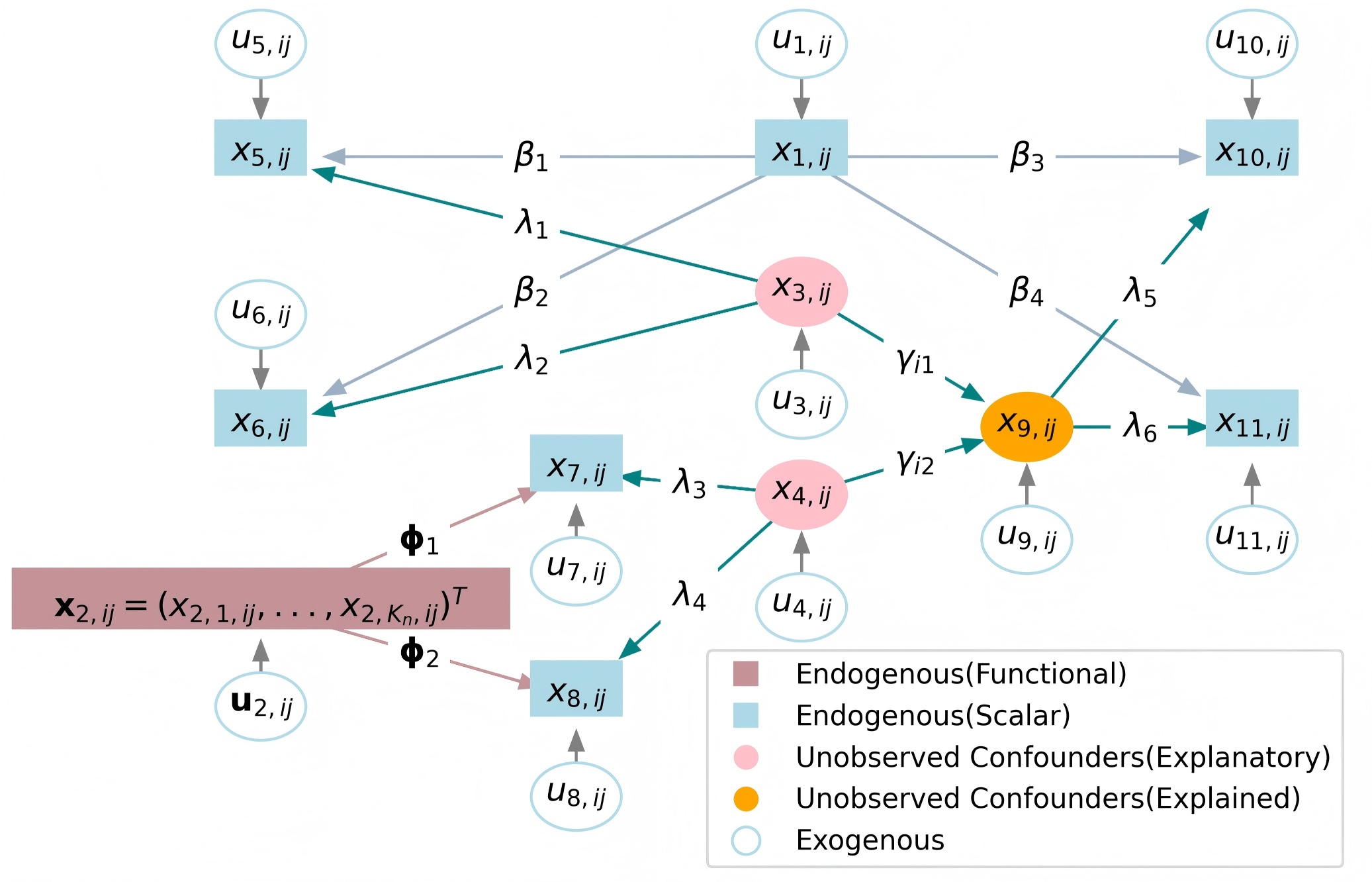}
\caption{PFST-DSCM with 11 exogenous and endogenous nodes (where nodes $
\textnormal{x}_{3,ij}$, $\textnormal{x}_{4,ij}$ and $\textnormal{x}_{9,ij}$ are unmeasured confounder nodes with spatial heterogeneity and temporal dependencies, and $\mathbf{x}_{2,ij}$ is functional node corresponding base expansion vector node)}
\label{DAG for PFST-DSCMs model}
\end{figure}

We considered a PFST-DSCM model comprising 11 exogenous and endogenous nodes, as illustrated in Fig. \ref{DAG for PFST-DSCMs model}. In this model, $\{\textnormal{x}_{3,ij}, \textnormal{x}_{4,ij}\}$ were set as unobserved explanatory variables, and $\textnormal{x}_{9,ij}$ was set as an unobserved response variable. We assumed that $\{\textnormal{x}_{3,ij}, \textnormal{x}_{4,ij}\}$ and $\textnormal{x}_{9,ij}$ exhibited pronounced spatial heterogeneity along with temporal dependence. Let $\textnormal{x}_{1,ij}$ represent the endogenous cause variable; $\mathbf{x}_{2,ij}$ represent the functional cause variable, $\{\textnormal{x}_{10,ij},\textnormal{x}_{11,ij}\}$ denote the outcome variables, and $\mathbf{x}_{\mathbb{B}}= \{\textnormal{x}_{5,ij}, \textnormal{x}_{6,ij},\textnormal{x}_{7,ij}, \textnormal{x}_{8,ij}\}$ constitute the backdoor adjustment sets. 

To systematically stress-test different methodological aspects, we generated data from eight distinct Data-Generating Processes (DGPs), creating a full-factorial design varying: i) Linearity (Linear vs. Nonlinear structural equations), ii) Noise Interaction (Additive vs. Non-additive) and iii) Noise Distribution (Gaussian vs. Heavy-tailed Student's $t$).   This allowed us to assess robustness across settings that were simple, challenging, and realistic. The specific explanations of the model names were shown in Table \ref{tab:model_acronyms} (Supplementary \ref{sec:Experiment details}).

\subsection{Simulation Settings}\label{sec:SSet}

\textbf{Baseline Models}. We benchmarked PFD-BDCM against a carefully chosen hierarchy of 10 alternative models, each designed to isolate the impact of a specific feature. 
First, the ``PFD-Oracle'' model served as an oracle, possessing access to the true structural equations. It represented the theoretical performance ceiling and provided an upper bound on achievable accuracy. The primary comparisons were against a series of ablation models derived from our own framework. ``PFD-DCM'' was PFD-BDCM without backdoor adjustment, isolating the value of incorporating this adjustment into the diffusion process. ``PF-DCM'' further ablated both backdoor adjustment and the dynamic modeling of confounders while retaining basis expansion for functional data. ``PF-BDCM'' included backdoor adjustment and basis expansion but treated unmeasured confounders as static, thereby isolating the specific benefit of modeling their dynamic nature. To validate our functional data integration strategy, we included several variants using mean-flattening instead of basis expansion: ``PFD-BDCM-FM'', ``PFD-DCM-FM'', ``PF-BDCM-FM'', and ``PF-DCM-FM''. These tested the impact of a simpler functional summary across different levels of our causal adjustment hierarchy. Finally, we compared against two traditional adjustment methods that condition on the same backdoor set but used non-generative models. ``PFD-BLR'' (Partially Functional Dynamic Backdoor Linear Regression) used linear regression to test the added value of the diffusion model's flexibility over a simple parametric model. ``PFD-BRF'' (Partially Functional Dynamic Backdoor Random Forest) used a random forest to test the benefit of our generative approach against a flexible, but non-generative, nonlinear model. This comprehensive hierarchy allowed us to precisely attribute performance gains to specific components of our proposed framework.

\textbf{Evaluation Protocol}. For each DGP and model, we evaluated performance on the three fundamental causal queries-observational (Obs.), interventional (Int.), and counterfactual (CF.)-using the Maximum Mean Discrepancy (MMD$^2$, \cite{mmd}) between the model-generated and true distributions, lower MMD$^2$ indicated better performance. We tested two sample sizes (J=6, n=30 and n=200) and two intervention types (uniform and quantile).  

Detailed simulation specifics, including the data-generating process, parameter settings, model configurations and training hyperparameters, were provided in Supplementary \ref{sec:Experiment details}.  

\subsection{Simulation Results}

\begin{figure}[!ht]
\centering    
    \includegraphics[width=1\linewidth]{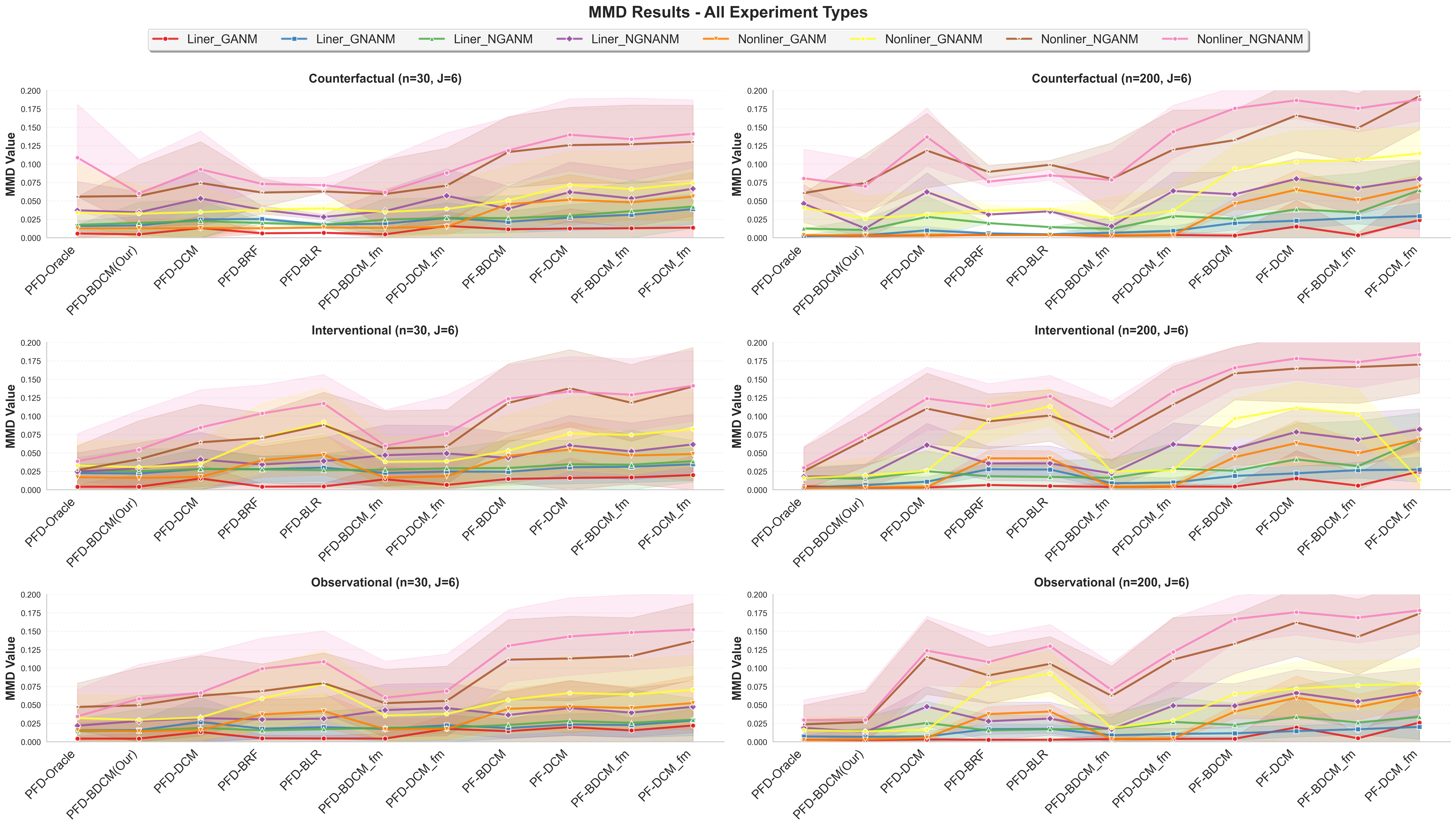}
     \caption{MMD Results for All Experiment Types-Quantile Intervention ($10\%,90\%$)}
    \label{fig:MMD_all-quantile}
\end{figure}

Complete observational, interventional, and counterfactual query results were shown in Fig. \ref{fig:MMD_all-quantile},  \ref{fig:MMD_all-uniform} and \ref{fig:Time_all}. Detailed results were provided in Supplementary \ref{sec:Experiment details} (Tables~\ref{sim: result1} and \ref{sim: result2} summarized the aggregated performance metrics
averaged over two independent random initializations). Kernel density estimates (Figs. \ref{fig:kde1}–\ref{fig:kde8}) were included in Supplementary \ref{sec:Experiment details}.

The aggregated results reveal a clear and differentiated pattern that directly supports our core methodological claims. First, we find compelling evidence for the superiority of our approach in settings where unmeasured confounders evolve over time. Under such conditions, PFD-BDCM consistently and substantially outperforms both PF-DCM, which ignores confounding altogether, and PF-BDCM, which incorrectly assumes that confounders remain static. 
The performance gains are especially pronounced for interventional and counterfactual queries, which are precisely the estimands most vulnerable to unadjusted bias. For example, in nonlinear, non-additive noise settings, the MMD$^2$ achieved by PFD-BDCM is often 20–40\% lower than that of PF-BDCM. These results provide direct empirical confirmation that explicitly modeling the temporal dependence and spatial heterogeneity of confounders is not merely a theoretical nicety but a practical imperative for reliable causal inference.

Turning to the role of backdoor adjustment within the diffusion process, the comparison between PFD-BDCM and its ablated counterpart PFD-DCM is instructive. Although both models share the same architectural backbone, PFD-DCM, which conditions only on parent nodes, performs markedly worse across all data-generating processes that involve unmeasured confounders. This difference underscores the effectiveness of our central innovation: integrating the backdoor adjustment set into the conditioning mechanism of the AC-DDIM, thereby actively correcting for confounding bias during the generative sampling phase.

We also observe clear support for our use of basis expansion to handle functional data. Across multiple comparisons, PFD-BDCM reliably outperforms its mean-flattening variant, PFD-BDCM-FM. The advantage is most evident in queries involving the functional cause node $\mathbf{x}_2$, where the basis expansion preserves information embedded in the shape of the curve-information that is lost when a simple mean is used instead. This finding validates our decision to treat functional variables as multi-dimensional nodes represented through basis coefficients.

Finally, our results speak to the robustness and scalability of the proposed framework. PFD-BDCM maintains strong performance even under heavy-tailed noise and in highly nonlinear regimes, conditions under which traditional methods such as PFD-BLR and PFD-BRF break down entirely. Moreover, as sample size increases, the model's performance improves and stabilizes, with MMD$^2$ converging steadily toward the Oracle benchmark. This behavior suggests  statistical consistency and reinforces the practical viability of our approach in real-world settings.

\section{Empirical Application}\label{sec:Application}

We employed the PFD-BDCM framework to examine spatio-temporal dynamic structural causal relationships among air pollutant indicators and their determinants. Our analysis encompassed 30 provincial-level administrative divisions across Chinese mainland during the period January 2015 to December 2020. The study integrates  China's provincial CO$_2$ emission inventories from the China Emission Accounts and Datasets (CEADs)~\citep{xu2024china} and emissions data for nine atmospheric pollutants from the Multi-scale Emission Inventory of China (MEIC)~\citep{geng2024efficacy} as response variables for air pollutant emissions.

Building upon prior research~\citep{ozcan2013nexus,
ZHU2021118323} and incorporating domain-specific characteristics of regional emissions, we systematically collected foundational determinants across ten conceptual dimensions. The comprehensive dataset comprised 118 indicator variables, through collinearity diagnostics and random forest-based feature selection, we retained 49 statistically robust indicators for subsequent modeling (detailed indicators shown in Supplementary \ref{real example} Table \ref{PEI}).  

We constructed the PFST-DSCM for the human-driven mechanisms of atmospheric pollutant emissions (HDM-APE) in China, as illustrated in Fig. \ref{PFST-DSCM-HDM-APE} (Supplementary \ref{real example} ). 
This model provided prior knowledge for systematically analyzing the impact of human activities on atmospheric emissions, wherein the structure of the causal graph and the estimation of its weight coefficients were derived through a partially functional structural equation model. Building on this, we utilize the data along with the structural relationship information to train a PFD-BDCM model specifically tailored for China's atmospheric pollutant emission indicators and their corresponding anthropogenic driving factors. To further validate our approach, we implemented a traditional Bayesian structural equation modeling baseline (PF-DSEM-Bays) for causal query estimation. 
Complete experimental specifications and supplementary materials were documented in Supplementary \ref{real example}, with observational query results presented in Table~\ref{PEI_result}.

\begin{table}[t]
\caption{Mean ($\times 10^{-2}$) $\pm$ standard deviation of MMD$^2$ of models compared to the true target distribution (Observation query)}\label{PEI_result}
\centering
\resizebox{1\textwidth}{!}{%
\begin{tabular}{c c c c | c c c c}
\toprule
\cmidrule(lr){1-4} \cmidrule(lr){5-8}
Variable & PFD-DCM & PFD-BDCM & PF-DSEM-Bays &
Variable & PFD-DCM & PFD-BDCM & PF-DSEM-Bays \\
\midrule
SO$_2$   & $\mathbf{1.130\pm0.008}$ & $1.155\pm0.008$ & $13.919\pm0.131$ &
PM$_{10}$ & $\mathbf{1.133\pm0.008}$ & $1.141\pm0.008$ & $11.961\pm0.156$ \\
NO$_x$   & $\mathbf{1.135\pm0.008}$ & $1.196\pm0.009$ & $17.854\pm0.157$ &
PM$_{2.5}$ & $1.168\pm0.008$ & $\mathbf{1.125\pm0.008}$ & $12.584\pm0.163$ \\
CO       & $\mathbf{1.121\pm0.008}$ & $1.154\pm0.009$ & $12.311\pm0.156$ &
BC       & $\mathbf{1.150\pm0.008}$ & $1.180\pm0.008$ & $16.745\pm0.186$ \\
VOC      & $1.208\pm0.008$ & $\mathbf{1.182\pm0.008}$ & $24.777\pm0.149$ &
OC       & $1.218\pm0.008$ & $\mathbf{1.192\pm0.009}$ & $21.109\pm0.182$ \\
NH$_3$   & $1.193\pm0.008$ & $\mathbf{1.142\pm0.008}$ & $36.627\pm0.151$ &
CO$_2$   & $1.146\pm0.008$ & $\mathbf{1.112\pm0.008}$ & $21.793\pm0.178$ \\
\bottomrule
\end{tabular}
}
\end{table}

Table~\ref{PEI_result} compares the performance of PFD-DCM, PFD-BDCM, and the baseline PF-DSEM-Bays on observational queries for ten atmospheric pollutants, including CO\textsubscript{2}. Both PFD-DCM and PFD-BDCM demonstrate significantly lower MMD$^2$ values (approximately $1.121-1.218 \times 10^{-2}$) across all pollutants compared to the baseline (ranging from $11.961-36.627 \times 10^{-2}$), confirming their superior capability in approximating the true observational data distribution. 
        This indicates that both models are highly effective, with their relative strengths varying slightly across different pollutants.

The robust observational performance of the model enables intervention analysis, allowing us to simulate the impact of policy measures or environmental changes on pollutant emissions. We conducted systematic intervention experiments on 44 drivers affecting atmospheric pollutant emissions at quantile levels of $0\%, 25\%, 50\%, 75\%$, and $100\%$ to quantitatively assess how varying intervention intensities influence emission levels. Intervention query results are provided in the supplementary materials (HDM-APE.zip). Here, we illustrate the intervention outcomes using a key social driver-``Familyscale-Small ($1-3$)'' (Fig. \ref{int1.})-as an example; the analysis process for other cases follows a similar approach.

We quantified the intervention effect using the difference between the mean residuals of the observational queries (true value $-$  observationally generated value) and the mean residuals of the interventional queries (true value $-$ interventionally generated value), defined as $\Delta$ = interventionally generated value $-$ observationally generated value, where $\Delta < 0$ indicated a reduction in emissions after the intervention, $\Delta>0$ indicated an increase. The results are shown in the last subplot of Fig. \ref{int1.}, with detailed interpretations in Supplementary \ref{real example}.

\section{Discussion}\label{sec:Concluding}

We have developed the Partially Functional Dynamic Backdoor Diffusion-based Causal Model (PFD-BDCM), a unified framework for causal inference with spatio-temporally dependent unmeasured confounding and mixed functional-scalar data. The methodology addresses the limitations of existing approaches that assume static confounders and lack principled functional data handling. By integrating a formal structural model (PFST-DSCM) with basis expansions and a backdoor-adjusted diffusion process, PFD-BDCM enables rigorous interventional and counterfactual queries from observational data under a known causal graph.

Our contributions are threefold. Methodologically, we synthesize functional data analysis, spatio-temporal statistics, and deep generative modeling. It explicitly models spatial heterogeneity and temporal dynamics in unmeasured confounders via CAR processes and represents functional variables through basis coefficients as standard graph nodes. Theoretically, we establish that causal effects are preserved under basis expansion with explicit error bounds and prove that counterfactual estimation error is controlled by reconstruction fidelity, with robustness to heavy-tailed noise and misspecification. Empirically, PFD-BDCM outperforms strong baselines across nonlinear, non-additive, and heavy-tailed settings in simulations, and yields actionable insights in a real-world air pollution study.

This framework can be extended in several directions. When the causal graph is only partially known, integrating structure learning algorithms could broaden applicability, though identifiability under dynamic confounding requires careful consideration. For functional data not well-approximated by a fixed basis, more flexible representations (e.g., neural networks or Gaussian processes) may be beneficial. 
An additional direction concerns settings where target outcomes are unavailable or populations differ substantially, necessitating causal transportability. Recent work on federated adaptive causal estimation (FACE) \citep{Han03072025} provides a principled framework for incorporating heterogeneous data from multiple sites while preserving privacy and communication efficiency; adapting such federated learning approaches could enable our framework to identify prevailing dynamic structures across sites when local outcomes are sparse or unobserved. 

Limitations include reliance on a known causal graph, potential inefficiency of fixed-basis expansions for complex functional data, and scalability challenges for large graphs. Future work may address these through structure learning integration, adaptive basis selection, and architectural innovations for amortized inference. Despite these limitations, PFD-BDCM provides a rigorous foundation for causal inference with complex spatio-temporal and functional data, with broad applicability in environmental science, epidemiology, and beyond.

 \noindent \textbf{Conflict of interest/Competing interests:}
The authors declare no competing interests. 
\noindent \textbf{ Acknowledgments:}
The authors would like to thank the referees, the Associate Editor, and the Editor for their constructive comments that improved the quality of this paper.

\appendix
\bibliographystyle{apalike}
  \bibliography{bibliography.bib}

\end{document}